# Angular separability of data clusters or network communities in geometrical space and its relevance to hyperbolic embedding

Alessandro Muscoloni[1] and Carlo Vittorio Cannistraci[1,2,*]

[1]Biomedical Cybernetics Group, Biotechnology Center (BIOTEC), Center for Molecular and Cellular Bioengineering (CMCB), Center for Systems Biology Dresden (CSBD), Department of Physics, Technische Universität Dresden, Tatzberg 47/49, 01307 Dresden, Germany
[2]Complex Systems Network Intelligence Group (CSNI), Tsinghua Laboratory of Brain and Intelligence (THBI), Tsinghua University, Beijing, China

*Corresponding author: Carlo Vittorio Cannistraci (kalokagathos.agon@gmail.com)

## Abstract

Analysis of 'big data' characterized by high-dimensionality such as word vectors and complex networks requires often their representation in a geometrical space by embedding. Recent developments in machine learning and network geometry have pointed out the hyperbolic space as a useful framework for the representation of this data derived by real complex physical systems. In the hyperbolic space, the radial coordinate of the nodes characterizes their hierarchy, whereas the angular distance between them represents their similarity. Several studies have highlighted the relationship between the angular coordinates of the nodes embedded in the hyperbolic space and the community metadata available. However, such analyses have been often limited to a visual or qualitative assessment. Here, we introduce the angular separation index (ASI), to quantitatively evaluate the separation of node network communities or data clusters over the angular coordinates of a geometrical space. ASI is particularly useful in the hyperbolic space - where it is extensively tested along this study - but can be used in general for any assessment of angular separation regardless of the adopted geometry. ASI is proposed together with an exact test statistic based on a uniformly random null model to assess the statistical significance of the separation. We show that ASI allows to discover two significant phenomena in network geometry. The first is that the increase of temperature in 2D hyperbolic network generative models, not only reduces the network clustering but also induces a 'dimensionality jump' of the network to dimensions higher than two. The second is that ASI can be successfully applied to detect the intrinsic dimensionality of network structures that grow in a hidden geometrical space.

# Introduction

Geometrical representation of data in an embedded space is crucial for big data science analysis [1]. Since many years, the Euclidean space has been considered of great importance for data representation and visualization in many domains of science, from image pattern recognition [2], [3] to computational biology [4], [5]. The hyperbolic space has recently attracted significant attention both for data [6] and network embedding [1], [7], [8] representation. Data samples or network nodes are represented in a hyperbolic disk where the radial coordinates indicate the network degree heterogeneity, and the angular coordinates their similarity. For a fixed number of dimensions, the hyperbolic space has more capacity than the Euclidean space, indeed the hyperbolic volume grows exponentially with its radius. In addition, hyperbolic geometry is better suited to embed data with tree-likeness or underlying hierarchical/heterogeneous structure [6].

After the seminal work published in 2008 by Serrano et al. [9] and by Boguñá et al. [10] on the relevant modelling information hidden in the metric space of complex networks, the last decade has seen an ever increasing interest on network geometry. The basic idea is that the nodes are located at certain geometrical coordinates, such that for each pair of nodes a distance is defined, and nodes that are closer to each other in the space are more likely to be connected in the network topology [9]. In particular, a generative model for networks in the hyperbolic space named popularity-similarity optimization (PSO) [11] has been shown to provide an explanation for the clustering, small-worldness, scale-freeness and even rich-clubness [1], [12] typically observed in many complex networked systems from the real world [13]. Since then, several popularity-similarity models have been developed for generating random geometric graphs in the hyperbolic space [9], [11], [14]–[19]. Notably, a generalization of the PSO model - named nonuniform PSO (nPSO) [17] – was introduced to grow also network community organization, which is a network feature significantly characterizing many real networks. Indeed, it was shown that the nPSO is better suited than the PSO as realistic benchmark for the evaluation of inference techniques [20]. On the other side, also algorithms for embedding real topologies in the hyperbolic space have been designed [1], [7], [8], [21]–[27], which can be adopted for example in community detection and link prediction applications [1], [22]–[24], to study the navigability of complex networks [10], [28], or in pioneering attempts to develop latent geometry network markers in order to detect the separation between two groups in different conditions [29].

A fundamental concept behind most of these models and algorithms is that the growth of complex networks is driven by a trade-off between two attractive forces: popularity and similarity [11]. The hyperbolic space provides a natural geometric framework for the representation of this trade-off [11]. Several studies have highlighted the relationship between the angular coordinates of the nodes and their community. For example, Wang et al. [23] proposed a method for hyperbolic embedding that exploits the community structure of the network for inferring the angular coordinates. Furthermore, different models have been developed for generating networks in the hyperbolic space with soft communities [15], [16] or a desired community structure [17]. Other research articles have compared the angular coordinates obtained from a hyperbolic embedding method with the community metadata available for the network nodes, such as the geographical locations in Internet network [7], [8], [21] or in airports network [25]. However, the relationship between the angular coordinates and the metadata is mostly presented in a qualitative manner, visually showing that nodes belonging to the same geographical location appear close to each other in the angular space. Recently, Faqeeh et al. [30] have dedicated a study on such relationship between hyperbolic embedding and community structure, and have introduced a measure of angular coherence in order to quantify the extent to which nodes within the same community have similar angular coordinates [30]. Such measure indicates the concentration or spread of the angular coordinates of a community, regardless of the angular coordinates of the other communities, and it does not take into account the community size (angular coordinates of a community might be less coherent simply because its size is large).

In this work, we introduce the angular separation index (ASI) to quantitatively evaluate the separation of node network communities or data clusters over the angular coordinates of a geometrical space. ASI is particularly useful in the hyperbolic space - where it is extensively tested along this study - but can be used in general for any assessment of angular separation regardless of the adopted geometry. ASI is proposed together with an exact test statistic based on a uniformly random null model to assess the statistical significance of the separation. Furthermore, this study will focus on offering examples of evaluation of community separation in complex networks embedded in the 2-dimensional (separation over the circle circumference) and 3-dimensional (separation over the sphere surface) hyperbolic space, which is recently a topic of utmost importance in the field of data science [1], [23], [30]. However, although here we will mainly offer examples in two or three dimensions, without loss of generality, the strategy adopted to design ASI is valid also for dimensions higher than three in any geometrical space in which angular separation is evaluated.

# Results and Discussion

**Rationale behind the definition of ASI**

The high-level idea of the ASI in 2D is the following. Suppose that for each node is given the polar coordinate in the circle circumference and the community membership. Let's focus on a certain community, since the procedure is analogous for the others.

1. The first step is to define the arc of the circumference that represents the zone occupied by the community, and therefore we detect the two nodes that correspond to the extremes of the arc, determining the boundaries of the community.
2. The next step is to count the number of mistakes for such community, given by the number of nodes belonging to other communities that are lying between the extremes of the community in consideration.
3. Then, the procedure is repeated separately for each community, and the total number of mistakes is obtained summing over the communities.
4. Such total value is then normalized by the random-worst-case-scenario value (which is obtained by considering the worst of R (number of iterations) evaluations on uniformly random angular node reshuffling, see Methods for details), in order to compute the final index, assuming values in the range [0, 1]. ASI = 0 when the total number of mistakes is as bad as the random-worst-case-scenario, and ASI = 1 when there are no mistakes at all. For the details on the procedure and the mathematical formula, please refer to the Methods section.

In order to determine whether the index obtained is significantly different from a random organization of the nodes over the circle circumference, a statistical test is also performed. The angular coordinates of the nodes are randomly reshuffled several times, the ASI is computed for each random reshuffling, and an empirical p-value is obtained as the proportion of random ASIs that are grater or equal than the observed ASI.

The high-level idea of the ASI in 3D is analogous to the 2D case, except for the computation of the mistakes. Indeed, while in 2D the zone occupied by the community is represented by an arc of the circumference, in 3D it corresponds to a portion of the sphere surface. In order to determine such portion we firstly detect the extremes of the community separately for the azimuth and elevation angles. The spherical area delimited by these extremes is then projected to a rectangular area and, over the points belonging to the community, we compute a convex hull, which represents the zone occupied by the community. The number of mistakes for such community is given by the number of nodes belonging to other communities that are lying inside the convex hull of the community in consideration.

**Examples of application of ASI for evaluations in community separation**

Figure 1 shows three examples of hyperbolic embeddings of networks in 2D and the related evaluations using ASI. In particular, in Figure 1A we have generated a synthetic network adopting the nPSO model, which is able to produce realistic networks in the hyperbolic disk with clustering, small-worldness, scale-freeness and a desired community organization [17]. The network has been generated with parameters $N = 100$ (network size), $m = 3$ (half of average degree), $T = 0.1$ (temperature, inversely related to the clustering coefficient), $C = 5$ (number of communities) and $\gamma = 3$ (power-law degree distribution exponent). The coordinates of the nodes have been inferred using the coalescent embedding algorithm [1]. The visualization of the embedding highlights a perfect separation of the communities over the angular coordinates, and this is reflected by the ASI = 1 (p-value < 0.001). In Figure 1B we have generated a nPSO network with the same parameters as the previous one, except for $T = 0.9$. A higher temperature corresponds to a network with lower clustering and with a higher mixing between the communities, therefore we would expect an embedding with higher mixing and worse angular separation. The visualization of the coordinates inferred using coalescent embedding [1] highlights that the communities are still grouped reasonably well, with some nodes misplaced far from the main clusters. This is indeed confirmed by the quantitative evaluation, reporting an ASI = 0.58, which is in the middle-high range and significantly different with respect to a random organization (p-value < 0.001). Figure 1C reports, for a proof of concept, the evaluation of the angular coordinates of Figure 1B after a random reshuffling. The visualization clearly shows that nodes belonging to the same community are randomly spread around the whole circumference, and the quantitative evaluation adequately reflects the situation with a very low ASI = 0.10 and without significant statistical difference from random (p-value = 0.172). A description of the nPSO model and coalescent embedding algorithm can be found in Supplementary Information.

Figure 2 shows a comparison of angular separation in 2D and 3D. The real network *opsahl_10* [31] is a social network between employees of a manufacturing company and the annotated communities indicate four different company locations. The network has been embedded adopting the coalescent embedding algorithm [1] both in the 2D and 3D hyperbolic space and the respective coordinates are visualized. While in 2D two communities are perfectly separated and the other two are slightly mixed, the 3D exploits the additional dimension to reach a perfect separation of the communities. This is quantitatively confirmed by the ASI evaluation, with a very high index for the 2D case (ASI = 0.91) and a perfect index for the 3D case (ASI = 1), statistically significant in both the scenarios (p-value < 0.001).

As a main application, we used the ASI to evaluate the 2D and 3D hyperbolic embedding performed adopting different mapping techniques both on real and synthetic networks. Table 1 reports the results for the 2D scenario on 8 real networks for which the community metadata were available. Each network has been embedded in the hyperbolic space using several state-of-the-art techniques: coalescent embedding [1], MCA [22] and HyperMap-CN [21]. For coalescent embedding and MCA different variants have been executed. The angular coordinates inferred by the methods have been evaluated according to the ASI, comparing them to the community annotation. The table reports the ASI of each embedding technique for the 8 real networks and the mean ASI over the dataset. Looking at the average performance, the analysis highlights that the coalescent embedding variants obtain a higher ASI than the MCA variants, which in turn surpass HyperMap-CN. This result is in line with the simulations reported in the study related to the MCA algorithm [22], showing that MCA displays an embedding accuracy that in general seems superior to HyperMap-CN and inferior to coalescent embedding. In particular, we noticed that the coalescent embedding RA2-ncISO obtains perfect angular separation for the *karate* and *opsahl_10* networks, as well as almost perfect for *opsahl_8* and *opsahl_11* networks. A description of the networks and details on the embedding methods can be found in Supplementary Information.

Table 2 shows the results of analogous simulations for the 3D scenario. We let notice that in this case only the coalescent embedding variants are reported, since the other techniques MCA and HyperMap-CN are able to embed only in 2D. The table highlights that the ASI values are overall higher than the 2D scenario, suggesting that the addition of the third dimension might be useful to obtain a higher angular separation, as previously commented in Figure 2.

Table 3 reports analogous results for both the 2D and 3D scenarios, but on synthetic nPSO networks. The networks have been generated using the nPSO model with parameters $N = [100, 1000]$, $m = [4, 8]$, $T = [0.1, 0.3, 0.5, 0.7]$, $C = [3, 6, 9]$ and $\gamma = [2, 3]$. For each combination of parameters, 100 networks have been generated. For each network the hyperbolic embedding methods have been executed and the ASI has been computed considering the coordinates inferred by the method and the ground-truth community information from the nPSO model. The table reports the mean ASI over the nPSO networks with $T = 0.1$, with $T = 0.7$, and over all the networks. The main result emerging from this analysis is a confirmation of the trends highlighted on real networks. The coalescent embedding variants obtain overall the best results, followed by MCA variants and finally by HyperMap-CN. Furthermore, the angular separation adopting 3D coordinates appears closer to the perfect separation with respect to the 2D scenario.

**Relevance and impact of ASI in understanding network geometry modelling**

The first important discovery that we achieve by means of ASI is reported in Table 3 and Figure 3, where we evaluate by ASI the node community separation in the 2D and 3D embedded hyperbolic space. Notice that the networks adopted for this test were generated in the 2D hyperbolic space by means of the nPSO model at different temperature levels. The temperature is an important parameter because (for increasing values) reduces the clustering by increasing the probability to create long range interactions between network nodes that do not reside in the same network neighbourhood. It is evident that for $T = 0.1$, the 3D embedding techniques perform better than the 2D ones. And, this difference significantly increases for higher temperature. This result is astonishing and remarkable at the same time. It is astonishing because we would expect that networks generated in a 2D space should have an intrinsic geometrical dimensionality that is 2D, and therefore should offer a better node discrimination if embedded in a 2D space; or alternatively, the node discrimination should not improve with the dimensionality. It is remarkable because we discover that the generation of long range interactions creates 'bridges' between far apart zones of the network and therefore generates a sort of 'dimensional tunnels' that increase the intrinsic dimensionality of the network. In practice, the long range interactions generated by a $T > 0$ put in contact zones that are not geometrically close in 2D, hence increasing the dimensionality of the network. To the best of our knowledge, it should be the first time that this phenomenon is explained. We need to report that the phenomenon was noticed in our previous publication [1], however at that stage we did not have any plausible and quantitative understanding of it. By means of the ASI, we provide solid evidences that the temperature increases the intrinsic dimensionality of the network generated by means of the PSO and nPSO model. The networks generated in the 2D space demonstrate already at $T = 0.1$ a 'dimensionality jump' that remarkably increases with the temperature (Fig. 3). Finally, ASI demonstrates its utility for detecting the intrinsic dimensionality of the data, which are associated in this case to a network structure.

To conclude, in this study we have introduced ASI, an index to quantitatively evaluate the separation of node network communities (or data clusters) over the angular coordinates of a geometrical space, which is particularly useful in the hyperbolic space, but can be used in general for any assessment of angular separation regardless of the adopted geometry. ASI is proposed together with an exact test statistic based on a uniformly random null model to assess the statistical significance of the separation. The proposed index covers both the 2D scenario with nodes arranged over a circle circumference and also the 3D case in which they are disposed on a sphere surface. However, the strategy on which the index is based is valid for any number of dimensions. Although here we test and comment ASI application for the evaluation of hyperbolic embedding techniques, it can be equivalently adopted for a different geometrical space such as Euclidean, since only the angular coordinates matter in the evaluation. With the design of this index we aim at overcoming the qualitative and visual assessments adopted in previous studies, providing a standard tool for a quantitative analysis of the relationship between the nodes distribution (over the angular coordinates) and the cluster/community labels in big data or complex network science.

## Methods

### Angular separation index (ASI)

The input of the algorithm for computing the angular separation index (ASI) is represented by the angular coordinates of the $N$ nodes in the 2D or 3D space and by the group (or community) memberships of the $N$ nodes. In the 2D case the angular coordinates are represented by the polar angle $\theta_{1...N} \in [0, 2\pi]$ of a polar coordinate system, whereas in the 3D case by the azimuth $\theta_{1...N} \in [0, 2\pi]$ and elevation $\varphi_{1...N} \in \left[-\frac{\pi}{2}, \frac{\pi}{2}\right]$ angles of a spherical coordinate system. Let's indicate with $G$ the number of groups.

The core of the algorithm for both the 2D and 3D cases is the following:

a) For each group $g = 1 ... G$, compute the number of mistakes $w_g$ in the angular arrangement. Computing the number of mistakes for a certain group consists in geometrically defining the extremes of that group and then counting how many nodes of other groups fall within those extremes. The procedure differs for the 2D and 3D cases and therefore will be described in details in the next sections.

b) Perform random reshufflings $r = 1 ... R$ (i.e. $R = 1000$) of the angular coordinates of the $N$ nodes.

c) For each random reshuffling $r = 1 ... R$ repeat step a), obtaining for each group $g = 1 ... G$ the number of mistakes $w_g^r$.

d) The ASI is obtained as:

$$ASI = 1 - \frac{\sum_g w_g}{\max_r \left(\sum_g w_g^r\right)}$$

Where $g = 1 ... G$ and $r = 1 ... R$. Note that $\max_r \left(\sum_g w_g^r\right)$ represents the computational worst case for the total amount of mistakes. The ASI therefore assumes values in the range [0, 1]: it is equal to 0 when the total amount of mistakes is identical to the worst case; the lower the amount of mistakes, the more it becomes close to 1; it is equal to 1 in the perfect scenario in which there are no mistakes at all.

e) In order to obtain an empirical null distribution of ASIs, for each random reshuffling the ASI is also computed:

$$ASI^r = 1 - \frac{\sum_g w_g^r}{\max_p \left(\sum_g w_g^p\right)}$$

Where $g = 1 ... G$, $r = 1 ... R$ and $p = 1 ... R$.

f) The empirical p-value for the statistical significance of the observed ASI is obtained as:

$$p = \frac{1 + \sum_r \delta(ASI^r \geq ASI)}{1 + R}$$

Where $r = 1 \ldots R$ and $\delta(x)$ is a function returning 1 if $x$ is true and 0 if $x$ is false.

**Computation of mistakes in 2D**

The main step in order to compute mistakes in 2D is to define the arc of the circumference that represents the zone occupied by the community, and therefore to detect the two nodes that correspond to the extremes of the arc, determining the boundaries of the community. In order to implement this procedure, the nodes are ranked according to the increasing angular coordinates $\theta_{1 \ldots N} \in [0, 2\pi]$ and ranks $r_{1 \ldots N} \in [1, N]$ are assigned. Let's consider a certain group $g \in [1, G]$, since the procedure is identical for all the groups. Let $r^g_{1 \ldots N^g}$ be the subset of the ranks $r_{1 \ldots N}$ for the $N^g$ nodes belonging to group $g$. The ranks $r^g_{1 \ldots N^g}$ are sorted in ascending order obtaining for the nodes of group $g$ the sorted ranks $s^g_{1 \ldots N^g} \in [1, N]$. Each rank in $s^g_{1 \ldots N^g}$ can be compared to the next one in order to compute the number of nodes belonging to different groups falling between the two nodes of group $g$ corresponding to those two ranks:

$$w_g(i) = s^g_{i+1} - s^g_i - 1 \qquad \text{for } i \in [1, N^g - 1]$$

Note that the sorted ranks $s^g_{1 \ldots N^g}$ should be interpreted circularly, therefore the last one is compared to the first one:

$$w_g(N^g) = N - s^g_{N^g} + s^g_1 - 1$$

The extremes of the group are defined by the two nodes with ranks that are (circularly) adjacent in $s^g_{1 \ldots N^g}$ and that have the highest number of nodes belonging to different groups falling between them. These specific nodes belonging to different groups are considered to fall "outside" the extremes, whereas all the other nodes of the network fall "inside". Nodes not belonging to group $g$ and falling inside the extremes of group $g$ represent the number of mistakes, which can be computed as:

$$w_g = \sum_{i=1 \ldots N^g} w_g(i) - \max_{i=1 \ldots N^g} w_g(i)$$

**Theoretical proof for the approximated value of the worst case scenario in 2D**

In the 3D space, the nodes cannot follow an order, therefore the mistakes are associated with the node localization on a surface. In the 2D space, the nodes are aligned on a line (circonference), therefore the mistakes are univocally associated with the node ordering on the 2D angular coordinate. This makes easy to derive theoretically the approximated value of the worst case scenario in 2D, which can be used to replace the $\max_r(\sum_g w_g^r)$ that represents a procedure to approximate by computation the worst case for the total amount of mistakes. This helps to reduce the time of computing of ASI in 2D space. The theoretical proof is simple: the total number of mistakes between the extremes of the group $g$ in the worst scenario in which the $N^g$ nodes are equidistantly arranged over the circumference is:

$$w_g^{worst} = \text{ceil}\left((N - N^g) * \frac{N^g - 1}{N^g}\right)$$

If the $N^g$ nodes of the group are equidistantly arranged in the circular ordering and the remaining $N - N^g$ nodes are distributed in the remaining positions of the ordering, then between two consecutive nodes of group $g$ there will be $\frac{N-N^g}{N^g}$ nodes. Out of this $N^g$ partitions of the $N - N^g$ nodes, one has to be considered out of the extremes of the group $g$, and the other $N^g - 1$ partitions are instead considered inside the extremes and therefore are counted as mistakes. Note that the ceil function is used to round for obtaining an integer number of mistakes.

Here we assume for each group that in the worst case its nodes are equidistantly arranged. This is correct as worst case when the single group is considered. However, considering all the groups together, a circular ordering in which all the groups have its nodes equidistantly arranged might be not always possible. It can be that some groups are equidistantly arranged and others are "almost" equidistantly arranged. If this happens the formula should give a number of mistakes slightly worse than the true worst configuration possible.

**Computation of mistakes in 3D**

As for the 2D case, let's consider a certain group $g \in [1, G]$, since the procedure is identical for all the groups. The first step consists in finding the extremes of the group, both regarding the azimuth $\theta$ and elevation $\varphi$ angles.

For the elevation, the extremes $\varphi_{ext1}^g$ and $\varphi_{ext2}^g$ are computed simply as the minimum and maximum among the values $\varphi_{1...N^g}^g \in \left[-\frac{\pi}{2}, \frac{\pi}{2}\right]$ of the group:

$$\varphi_{ext1}^g = \min_{i=1...N^g} \varphi_i^g$$

$$\varphi_{ext2}^g = \max_{i=1...N^g} \varphi_i^g$$

For the azimuth, the extremes are computed with the exact same procedure described for the polar angle $\theta$ in the 2D case, explained in the previous section. Let $\theta_{ext1}^g$ and $\theta_{ext2}^g$ be the azimuth angles of the two extremes, with $\theta_{ext1}^g < \theta_{ext2}^g$. Note that if the index $i \in [1, N^g]$ to maximize $w_g(i)$ (see previous section) is $i = N^g$, then the azimuth values $\theta^*$ inside the two extremes will satisfy $\theta^g_{ext1} < \theta^* < \theta^g_{ext2}$ (set flag = 0), instead if $i < N^g$ then the azimuth values within the two extremes will satisfy $\theta^g_{ext2} < \theta^* < 2\pi$ OR $0 < \theta^* < \theta^g_{ext1}$ (set flag = 1). After having computed the extremes of the group $g$, the next step is to detect all the nodes falling inside the spherical surface delimited by the azimuth and elevation extremes. Such nodes would have angular coordinates $\theta^*, \varphi^*$ that satisfy:

$$\varphi^g_{ext1} < \varphi^* < \varphi^g_{ext2}$$

$$AND$$

$$\begin{pmatrix} (flag = 0 \text{ AND } \theta^g_{ext1} < \theta^* < \theta^g_{ext2}) \\ OR \\ (flag = 1 \text{ AND } (\theta^* < \theta^g_{ext1} \text{ OR } \theta^* > \theta^g_{ext2})) \end{pmatrix}$$

After having detected such nodes, their azimuth and elevation coordinates $\theta^*, \varphi^*$ in the spherical surface are mapped to Cartesian coordinates $x^*, y^*$ of a rectangular 2D area.

$$y^* = \varphi^* - \varphi^g_{ext1}$$

$$x^* = \theta^* - \theta^g_{ext1} \quad \text{if flag} = 0$$

$$x^* = \text{modulo}(\theta^* + (2\pi - \theta^g_{ext2}), 2\pi) \quad \text{if flag} = 1$$

The next step is to compute the convex hull of the points of the group $g$ from their Cartesian coordinates $x^{*,g}, y^{*,g}$. The convex hull delimits the area of the group $g$, therefore for each point that has been mapped to Cartesian coordinates and that does not belong to group $g$, it can be tested whether it falls inside or outside the polygon. The number of nodes that do not belong to group $g$ falling inside the polygon represents the number of mistakes $w_g$.

**Code availability**

The MATLAB code for computing the angular separation index (ASI) is publicly available at the GitHub repository:

https://github.com/biomedical-cybernetics/coalescent_embedding/tree/master/visualization_and_evaluation/angular_separation_index

**Hardware and software**

MATLAB code has been used for all the simulations, carried out partly on a workstation under Windows 8.1 Pro with 512 GB of RAM and 2 Intel(R) Xenon(R) CPU E5-2687W v3 processors with 3.10 GHz, and partly on the ZIH-Cluster Taurus of the TU Dresden.

**Funding**

Work in the CVC laboratory was supported by the independent research group leader starting grant of the Technische Universität Dresden.

**Acknowledgements**

We thank the BIOTEC System Administrators for their IT support, Gloria Marchesi and Michelle Weichlein for the administrative assistance and the Centre for Information Services and High Performance Computing (ZIH) of the TUD.

**Author contributions**

CVC conceived the idea of angular separation index and the content of the study. Both the authors contributed in designing the angular separation index, the numerical experiments and the items of the draft. CVC derived the theoretical proof for the approximated value of the worst case scenario in 2D. AM implemented the code, performed the computational analysis and realized the items of the draft. Both the authors analyzed and interpreted the results. AM wrote the draft of the article according to CVC suggestions, CVC corrected and improved it. CVC planned, directed and supervised the study.

**Competing interests**

The authors declare no competing financial interests.

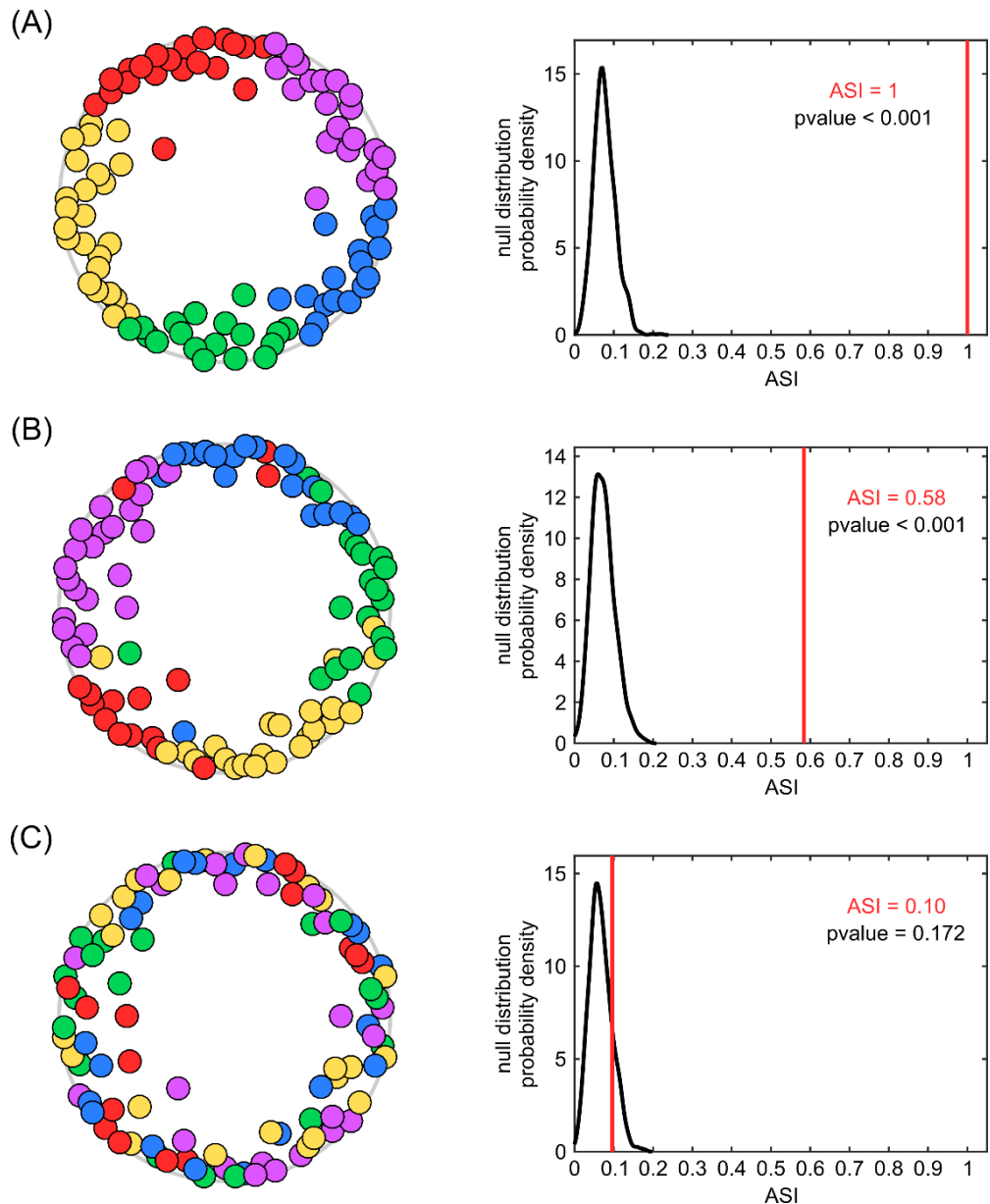

**Figure 1. Angular separation in 2D with statistical test.**
The left panels show examples of 2D hyperbolic embeddings of synthetic networks generated using the nPSO model. **(A)** The nPSO network has been generated with parameters $N = 100$ (network size), $m = 3$ (half of average degree), $T = 0.1$ (temperature, inversely related to the clustering coefficient), $C = 5$ (number of communities) and $\gamma = 3$ (power-law degree distribution exponent). The embedded coordinates have been inferred using the coalescent embedding method RA2-ncISO-EA. The 5 ground-truth communities are highlighted with different colours. **(B)** The nPSO network has been generated with the same parameters as in (A), except for $T = 0.9$. The embedded coordinates have been inferred using the coalescent embedding method RA2-ncISO-EA. **(C)** The embedded coordinates correspond to the ones in (B) after a random reshuffling. The right panels represent the statistical test for the ASI evaluation and show the observed ASI (in red) compared to the null distribution of ASIs (in black), reporting the related p-value.

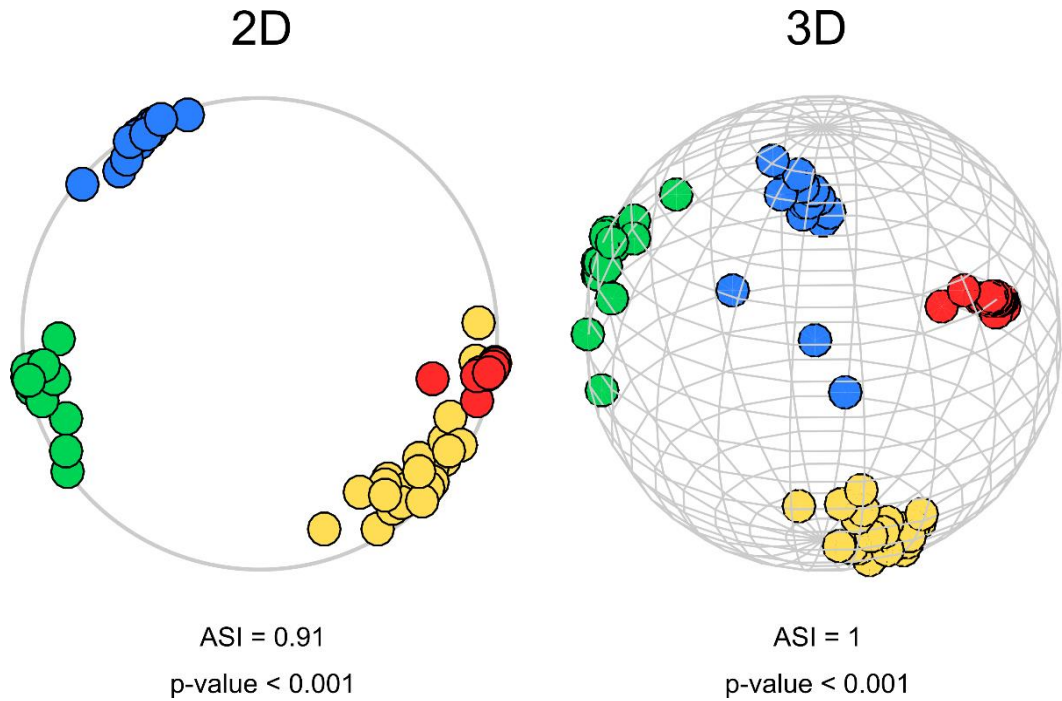

**Figure 2. ASI improvement in 3D with respect to 2D.**
The figure shows the hyperbolic embedding of the *opsahl_10* network using the coalescent embedding method RA1-ISO both in the 2D hyperbolic disk (left) and in the 3D hyperbolic sphere (right). The 4 ground-truth communities are highlighted with different colours. At the bottom of each panel the ASI and the related p-value of the statistical test are reported. The figure provides an example in which the addition of the third dimension of embedding improves the angular separation of the communities, leading to a perfect segregation.

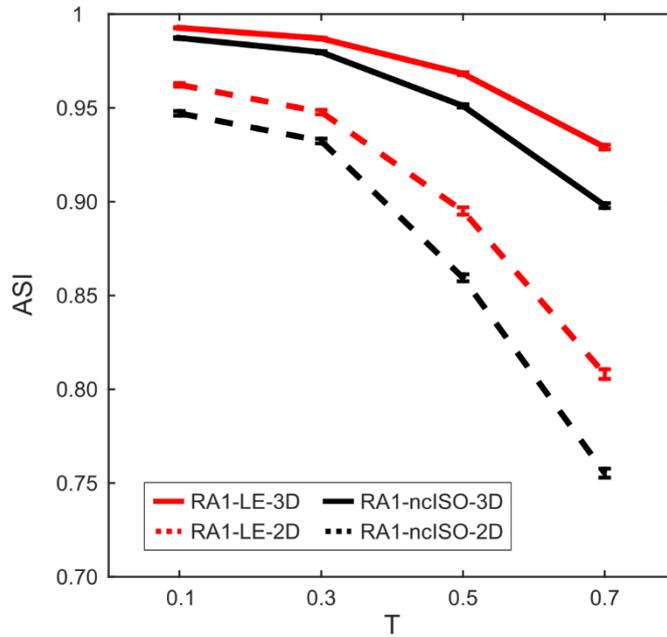

**Figure 3. ASI evaluation of community separation in nPSO networks for increasing temperatures.** Synthetic networks have been generated using the nPSO model with parameters $N = [100, 1000]$ (network size), $m = [4, 8]$ (half of average degree), $T = [0.1, 0.3, 0.5, 0.7]$ (temperature, inversely related to the clustering coefficient), $C = [3, 6, 9]$ (number of communities) and $\gamma = [2, 3]$ (power-law degree distribution exponent). For each combination of parameters, 100 networks have been generated. For each network the hyperbolic embedding methods have been executed and the ASI has been computed considering the coordinates inferred by the method and the ground-truth community information from the nPSO model. The plot reports the mean ASI and standard error for each value of $T$, averaging over all the other parameters.

|  | mean ASI | karate | opsahl 8 | opsahl 9 | opsahl 10 | opsahl 11 | polbooks | football | polblogs |
|---|---|---|---|---|---|---|---|---|---|
| RA2-ncISO | 0.78 | 1.00 | 0.93 | 0.80 | 1.00 | 0.99 | 0.60 | 0.80 | 0.15 |
| RA2-ISO | 0.77 | 0.90 | 0.89 | 0.79 | 0.99 | 1.00 | 0.61 | 0.81 | 0.17 |
| RA2-LE | 0.77 | 0.93 | 0.93 | 0.86 | 1.00 | 0.86 | 0.56 | 0.81 | 0.19 |
| RA1-ncMCE | 0.77 | 0.93 | 0.90 | 0.81 | 1.00 | 0.99 | 0.63 | 0.84 | 0.02 |
| RA1-LE | 0.75 | 0.93 | 0.92 | 0.87 | 1.00 | 0.87 | 0.56 | 0.82 | 0.01 |
| RA2-ncMCE | 0.74 | 0.93 | 0.93 | 0.68 | 1.00 | 1.00 | 0.41 | 0.82 | 0.15 |
| RA2-MCE | 0.74 | 0.87 | 0.93 | 0.69 | 1.00 | 0.97 | 0.45 | 0.84 | 0.16 |
| RA1-ISO | 0.73 | 0.73 | 0.74 | 0.79 | 0.91 | 1.00 | 0.57 | 0.81 | 0.26 |
| RA1-ncISO | 0.70 | 0.93 | 0.70 | 0.76 | 0.84 | 0.88 | 0.53 | 0.81 | 0.17 |
| RA1-MCE | 0.70 | 0.67 | 0.88 | 0.80 | 0.92 | 0.89 | 0.60 | 0.83 | 0.02 |
| RA2-MCA2-RAA | 0.68 | 0.80 | 0.80 | 0.63 | 0.84 | 0.91 | 0.47 | 0.84 | 0.14 |
| RA1-MCA2-RAA | 0.67 | 0.86 | 0.77 | 0.47 | 0.92 | 0.81 | 0.49 | 0.84 | 0.20 |
| RA2-MCA1-RAA | 0.65 | 0.60 | 0.64 | 0.51 | 0.92 | 1.00 | 0.46 | 0.89 | 0.17 |
| RA1-MCA1-RAA | 0.62 | 0.45 | 0.64 | 0.54 | 0.92 | 0.91 | 0.45 | 0.87 | 0.19 |
| HyperMap-CN | 0.60 | 0.83 | 0.80 | 0.78 | 0.64 | 0.79 | 0.33 | 0.52 | 0.10 |

**Table 1. ASI evaluation of 2D embedding in real networks.**
For each real network the 2D hyperbolic embedding methods have been executed and the ASI has been computed considering the coordinates inferred by the method and the ground-truth community information from the metadata. The table reports the ASI of each embedding technique for the 8 real networks and the mean ASI over the dataset. The methods are ranked by mean performance. A description of the networks and embedding methods can be found in Supplementary Information.

|  | mean ASI | karate | opsahl 8 | opsahl 9 | opsahl 10 | opsahl 11 | polbooks | football | polblogs |
| --- | --- | --- | --- | --- | --- | --- | --- | --- | --- |
| RA2-LE | 0.86 | 0.96 | 0.98 | 0.96 | 1.00 | 1.00 | 0.66 | 0.97 | 0.38 |
| RA2-ncISO | 0.86 | 1.00 | 0.94 | 0.94 | 1.00 | 1.00 | 0.67 | 0.93 | 0.36 |
| RA1-ncISO | 0.85 | 0.95 | 0.87 | 0.94 | 0.99 | 1.00 | 0.64 | 0.93 | 0.48 |
| RA1-ISO | 0.84 | 0.96 | 0.85 | 0.95 | 1.00 | 1.00 | 0.64 | 0.92 | 0.41 |
| RA2-ISO | 0.83 | 0.96 | 0.86 | 0.93 | 1.00 | 1.00 | 0.67 | 0.91 | 0.29 |
| RA1-LE | 0.82 | 1.00 | 0.98 | 0.96 | 1.00 | 1.00 | 0.64 | 0.96 | 0.03 |

**Table 2. ASI evaluation of 3D embedding in real networks.**
For each real network the 3D hyperbolic embedding methods have been executed and the ASI has been computed considering the coordinates inferred by the method and the ground-truth community information from the metadata. The table reports the ASI of each embedding technique for the 8 real networks and the mean ASI over the dataset. The methods are ranked by mean performance. A description of the networks and embedding methods can be found in Supplementary Information.

| 2D | $T = 0.1$ ASI | $T = 0.7$ ASI | mean ASI | 3D | $T = 0.1$ ASI | $T = 0.7$ ASI | mean ASI |
|---|---|---|---|---|---|---|---|
| RA1-LE | 0.96 | 0.81 | 0.90 | RA1-LE | 0.99 | 0.93 | 0.97 |
| RA2-LE | 0.95 | 0.81 | 0.90 | RA2-LE | 0.99 | 0.93 | 0.97 |
| RA1-ncISO | 0.95 | 0.76 | 0.87 | RA1-ncISO | 0.99 | 0.90 | 0.95 |
| RA1-ISO | 0.95 | 0.75 | 0.87 | RA1-ISO | 0.99 | 0.90 | 0.95 |
| RA2-ncISO | 0.94 | 0.74 | 0.86 | RA2-ncISO | 0.98 | 0.90 | 0.95 |
| RA2-ISO | 0.94 | 0.73 | 0.85 | RA2-ISO | 0.98 | 0.89 | 0.95 |
| RA2-ncMCE | 0.92 | 0.64 | 0.82 | | | | |
| RA2-MCE | 0.90 | 0.62 | 0.80 | | | | |
| RA1-ncMCE | 0.86 | 0.61 | 0.77 | | | | |
| RA2-MCA2-RAA | 0.88 | 0.56 | 0.75 | | | | |
| RA1-MCE | 0.81 | 0.57 | 0.72 | | | | |
| RA2-MCA1-RAA | 0.85 | 0.48 | 0.69 | | | | |
| RA1-MCA2-RAA | 0.81 | 0.51 | 0.67 | | | | |
| HyperMap-CN | 0.79 | 0.47 | 0.63 | | | | |
| RA1-MCA1-RAA | 0.76 | 0.40 | 0.60 | | | | |

**Table 3. ASI evaluation of 2D and 3D embedding in nPSO networks.**
Synthetic networks have been generated using the nPSO model with parameters $N = [100, 1000]$ (network size), $m = [4, 8]$ (half of average degree), $T = [0.1, 0.3, 0.5, 0.7]$ (temperature, inversely related to the clustering coefficient), $C = [3, 6, 9]$ (number of communities) and $\gamma = [2, 3]$ (power-law degree distribution exponent). For each combination of parameters, 100 networks have been generated. For each network the hyperbolic embedding methods have been executed and the ASI has been computed considering the coordinates inferred by the method and the ground-truth community information from the nPSO model. The table reports the mean ASI over the nPSO networks with $T = 0.1$, with $T = 0.7$, and over all the networks, both for 2D (left columns) and 3D (right columns) embedding techniques. The methods are ranked by mean performance over all the networks. A description of the networks and embedding methods can be found in Supplementary Information.

# SUPPLEMENTARY INFORMATION

# Angular separability of data clusters or network communities in geometrical space and its relevance to hyperbolic embedding

Alessandro Muscoloni[1] and Carlo Vittorio Cannistraci[1,2,*]

[1]Biomedical Cybernetics Group, Biotechnology Center (BIOTEC), Center for Molecular and Cellular Bioengineering (CMCB), Center for Systems Biology Dresden (CSBD), Department of Physics, Technische Universität Dresden, Tatzberg 47/49, 01307 Dresden, Germany
[2]Complex Systems Network Intelligence Group (CSNI), Tsinghua Laboratory of Brain and Intelligence (THBI), Tsinghua University, Beijing, China

*Corresponding author: Carlo Vittorio Cannistraci (kalokagathos.agon@gmail.com)

# 1. Datasets

## 1.1 Real networks

The community detection methods have been tested on 8 real networks, which represent differing systems: Karate; Opsahl_8; Opsahl_9; Opsahl_10; Opsahl_11; Polbooks; Football; Polblogs. The networks have been transformed into undirected, unweighted, without self-loops and only the largest connected component has been considered. The information of some basic statistics are available in Suppl. Table 1. $N$ is the number of nodes. $E$ is the number of edges. The parameter $m$ refers to half of the average node degree and it is also equal to the ratio $E/N$. $Cl$ is the average clustering coefficient, computed for each node as the number of links between its neighbours over the number of possible links [1]. The parameter $\gamma$ is the exponent of the power-law degree distribution, fitted from the observed degree sequence using the maximum likelihood procedure developed by Clauset et al. [2] and released at http://tuvalu.santafe.edu/~aaronc/powerlaws/. $C$ is the number of ground-truth communities.

The first network is about the Zachary's Karate Club [3], it represents the friendship between the members of a university karate club in US. The communities are formed by a split of the club into two parts, each following one trainer.

The networks from the second to the fifth are intra-organisational networks from [4] and can be downloaded at https://toreopsahl.com/datasets/#Cross_Parker. Opsahl_8 and Opsahl_9 come from a consulting company and nodes represent employees. In Opsahl_8 employees were asked to indicate how often they have turned to a co-worker for work-related information in the past, where the answers range from: 0 - I don't know that person; 1 - Never; 2 - Seldom; 3 - Sometimes; 4 - Often; 5 - Very often. Directions were ignored. The data was turned into an unweighted network by setting a link only between employees that have at least asked for information seldom (2).

In the Opsahl_9 network, the same employees were asked to indicate how valuable the information they gained from their co-worker was. They were asked to show how strongly they agree or disagree with the following statement: "In general, this person has expertise in areas that are important in the kind of work I do." The weights in this network are also based on the following scale: 0 - Do Not Know This Person; 1 - Strongly Disagree; 2 - Disagree; 3 - Neutral; 4 - Agree; 5 - Strongly Agree. We set a link if there was an agreement (4) or strong agreement (5). Directions were ignored.

The Opsahl_10 and Opsahl_11 networks come from the research team of a manufacturing company and nodes represent employees. The annotated communities indicate the company locations (Paris, Frankfurt, Warsaw and Geneva). For Opsahl_10 the researchers were asked to indicate the extent to which their co-workers provide them with information they use to accomplish their work. The answers were on the following scale: 0 – I do not know this person / I never met this person; 1 – Very infrequently; 2 – Infrequently; 3 – Somewhat frequently; 4 – Frequently; 5 – Very frequently. We set an undirected link when there was at least a weight of 4.

For Opsahl_11 the employees were asked about their awareness of each other's knowledge ("I understand this person's knowledge and skills. This does not necessarily mean that I have these skills and am knowledgeable in these domains, but I understand what skills this person has and domains they are knowledgeable in."). The weighting was on the scale: 0 – I do not know this person / I have never met this person; 1 – Strongly disagree; 2 – Disagree; 3 – Somewhat disagree; 4 – Somewhat agree; 5 – Agree; 6 – Strongly agree. We set a link when there was at least a 4, ignoring directions.

The Polbooks network represents frequent co-purchases of books concerning US politics on amazon.com. Ground-truth communities are given by the political orientation of the books as either conservative, neutral or liberal. The network is unpublished but can be downloaded at http://www-personal.umich.edu/~mejn/netdata/, as well as with the Karate, Football and Polblogs networks.

The Football network [5] presents games between division IA colleges during regular season fall 2000. Ground-truth communities are the conferences that each team belongs to.

The Polblogs [6] network consists of links between blogs about the politics in the 2004 US presidential election. The ground-truth communities represent the political opinions of the blogs (right/conservative and left/liberal). We let notice that most of this Methods section is equivalent to an analogous Methods section present in other studies of the authors [7], [8].

### 1.2 Synthetic networks generated by the nPSO model

The Popularity-Similarity-Optimization (PSO) model [9] is a generative network model recently introduced in order to describe how random geometric graphs grow in the hyperbolic space. In this model the networks evolve optimizing a trade-off between node popularity, abstracted by the radial coordinate, and similarity, represented by the angular distance. The PSO model can reproduce many structural properties of real networks: clustering, small-worldness (concurrent low characteristic path length and high clustering), node degree heterogeneity with power-law degree distribution and rich-clubness. However, being the nodes uniformly distributed over the angular coordinate, the model lacks a non-trivial community structure.

The nonuniform PSO (nPSO) model [10], [11] is a variation of the PSO model that exploits a nonuniform distribution of nodes over the angular coordinate in order to generate networks characterized by communities, with the possibility to tune their number, size and mixing property. We adopted a Gaussian mixture distribution of angular coordinates, with communities that emerge in correspondence of the different Gaussians, and the parameter setting suggested in the original study [10], [11]. Given the number of components $C$, they have means equidistantly arranged over the angular space, $\mu_i = \frac{2\pi}{C} \cdot (i - 1)$, the same standard deviation fixed to 1/6 of the distance between two adjacent means, $\sigma_i = \frac{1}{6} \cdot \frac{2\pi}{C}$, and equal mixing proportions, $\rho_i = \frac{1}{C}$ ($i = 1 \dots C$). The community memberships are assigned considering for each node the component whose mean is the closest in the angular space. The other parameters of the model are the number of nodes $N$, half of the average node degree $m$, the network temperature $T$ (inversely related to the clustering) and the exponent $\gamma$ of the power-law degree distribution. Given the parameters ($N$, $m$, $T$, $\gamma$, $C$), for details on the generative procedure please refer to the original study [10], [11]. The MATLAB code is publicly available at the GitHub repository: https://github.com/biomedical-cybernetics/nPSO_model.

## 2. Hyperbolic embedding
### 2.1 Coalescent embedding

The expression coalescent embedding refers to a topological-based machine learning class of algorithms that exploits nonlinear unsupervised dimensionality reduction to infer the nodes angular coordinates in the hyperbolic space [7]. The techniques are able to perform a fast and accurate mapping of a network in the 2D hyperbolic disk and in the 3D hyperbolic sphere.

The first step of the algorithm for a 2D embedding consists in pre-weighting the network in order to suggest geometrical distances between connected nodes, since it has been shown that improves the mapping accuracy [7]. Two topological-based pre-weighting rules have been proposed, repulsion-attraction (RA) and edge-betweenness-centrality (EBC), respectively using local (RA) and global (EBC) topological information [7]. In this work, we have adopted the following pre-weighting rules:

$$x_{ij}^{RA1} = \frac{1 + e_i + e_j}{1 + CN_{ij}}$$

$$x_{ij}^{RA2} = \frac{1 + e_i + e_j + e_i e_j}{1 + CN_{ij}}$$

Given the weighted network, the second step consists in performing the nonlinear dimensionality reduction. Two different kinds of machine learning approaches can be used, manifold-based (LE, ISO, ncISO) or Minimum-Curvilinearity-based (MCE, ncMCE). The details about which dimensions of the embedding should be considered are provided in the original publication [7].

In order to assign the angular coordinates in the 2D embedding space, either a circular adjustment or an equidistant angular adjustment (EA) can be considered. The circular adjustment for the manifold-based approaches consists in exploiting directly the polar coordinates of the 2D reduced space, whereas for the Minimum-Curvilinearity-based it consists in rearranging the node points on the circumference following the same ordering of the 1D reduced space and proportionally preserving the distances. Using the equidistant angular adjustment, instead, the nodes are equidistantly arranged on the circumference, which might help to correct for short-range angular noise present in the embedding [7]. However, note that the EA does not affect the angular separation.

In order to assign the radial coordinates, nodes are sorted according to descending degree and the radial coordinate of the $i$-th node in the set is computed according to:

$$r_i = \frac{2}{\zeta}[\beta \ln i + (1 - \beta) \ln N] \quad i = 1,2, \dots, N$$

$N$ number of nodes; $\zeta = \sqrt{-K}$, we set $\zeta = 1$; $K$ curvature of the hyperbolic space; $\beta = \frac{1}{\gamma - 1}$ popularity fading parameter; $\gamma$ exponent of power-law degree distribution.

The exponent $\gamma$ of the power-law degree distribution has been fitted using the MATLAB script '*plfit.m*', according to a procedure described by Clauset et al. [2] and released at http://www.santafe.edu/~aaronc/powerlaws/. If a network has $\gamma < 2 \rightarrow \beta > 1$, which is out of the domain $\beta \in (0, 1]$ imposed by the PSO model, a few of the highest degree nodes should obtain a radial coordinate that indicates a popularity higher than the maximum popularity allowed ($r = 0$), but since it is not possible due to the previous equation, the radial coordinate degenerates and becomes negative. Hence, for these nodes we set $r = 0$. The code is available at https://github.com/biomedical-cybernetics/coalescent_embedding. We let notice that most of this Methods section is equivalent to an analogous Methods section present in other studies of the authors [12].

## 2.2 Minimum Curvilinear Automaton (MCA)

MCA are network automata growing according to a strategy named minimum curvilinearity [13]. The idea of minimum curvilinearity (MC) is that the hidden geometry of complex networks that are in particular sufficiently sparse, clustered, small-world and heterogeneous can be efficiently navigated using the minimum spanning tree (MST), which is a greedy navigator. During each step of its greedy growing process (adopting the Prim's algorithm [14]), the MST sequentially attaches the node most similar (less distant) to its own cohort. Since the nodes angular coordinates in the hyperbolic disk actually represent an ordering of their similarities, they can be efficiently approximated by the visited node sequence of the MST.

The first step of the algorithm it the pre-weighting of the network according to the repulsion-attraction rules described for the coalescent embedding algorithm in the previous section (RA1 and RA2). As second step, the Prim's algorithm [14] is executed over the pre-weighted network, initializing the MST with the highest degree node. At each next iteration $t = 2 \dots N$, it finds the edge of minimum weight between nodes already in the tree and nodes not yet in the tree. Such edge, and the related node not yet in the tree, are attached to the tree. The sequence in which the nodes are introduced in the growing MST represents a circular ordering of their similarities. In a first variant (MCA1) the sequence grows only in one direction, in a second variant (MCA2) the sequence can grow in both the directions [13]. Then, the nodes are arranged over the circumference of the disk according to the circular ordering obtained at the previous step. In this work the angular distances between adjacent nodes have been fixed using the RAA adjustment [13], although such distances does not affect the angular separation. The radial coordinates are assigned as described for the coalescent embedding algorithm in the previous section. The code has been implemented by the authors in MATLAB. We let notice that most of this Methods section is equivalent to an analogous Methods section present in other studies of the authors [12].

## 2.3 HyperMap-CN

HyperMap [15] is a method to map a network into the hyperbolic space based on Maximum Likelihood Estimation (MLE). For sake of clarity, the first algorithm for MLE-based network embedding in the hyperbolic space is not HyperMap, but to the best of our knowledge is the algorithm proposed by Boguñá et al. in [16]. HyperMap is basically an extension of that method applied to the PSO model [9]. It replays the hyperbolic growth of the network and at each time step $i$ it finds the coordinates of the added node $i$ by maximizing the likelihood that the network was produced by the E-PSO model [15]. According to the MLE procedure, the nodes are added in decreasing order of degree. The radial coordinates depend on the time step $i$ and on the exponent $\gamma$ of the power-law degree distribution. The angular coordinates, instead, are assigned by maximizing a likelihood function $L_{i,L}$, with the aim of mapping connected nodes at a low hyperbolic distance and disconnected nodes at a high hyperbolic distance. The maximization is done by numerically trying different angular coordinates in steps of $2\pi/N$ and choosing the one that leads to the biggest $L_{i,L}$.

HyperMap-CN [17] is a further development of HyperMap, where the inference of the angular coordinates is not performed anymore maximizing the likelihood $L_{i,L}$, based on the connections and disconnections of the nodes, but using another local likelihood $L_{i,CN}$, based on the number of common neighbours between each node $i$ and the previous nodes $j < i$ at final time. Here the hybrid model has been used, a variant of the method in which the likelihood $L_{i,CN}$ is only adopted for the high degree nodes and $L_{i,L}$ for the others, yielding a shorter running time. Furthermore, a speed-up heuristic and corrections steps can be applied. The speed-up can be achieved by getting an initial estimate of the angular coordinate of a node $i$ only considering the previous nodes $j < i$ that are $i$'s neighbours, the maximum likelihood estimation is then performed only looking at an interval around this initial estimate. Correction steps can be used at predefined times $i$: each existing node $j < i$ is visited and with the knowledge of the rest of the coordinates the angle of $j$ is updated to the value that maximizes the likelihood $L_{j,L}$. The C++ implementation of the method has been released by the authors at the website https://bitbucket.org/dk-lab/2015_code_hypermap. In our simulations, neither correction steps nor speed-up heuristic have been used. The input parameter $\gamma$ has been fitted as described for the coalescent embedding method. Based on the assumption that the clustering coefficient decreases almost linearly with the network temperature, until it is 0 for $T = 1$ [9], the following procedure has been proposed [18] in order to choose the input parameter $T$ (temperature): (1) ten PSO synthetic networks are generated with $T = 0$ and the same parameters $N$, $m$ and $\gamma$ as the given network; (2) the clustering coefficient of the ten networks is averaged and used as y-intercept, while the point ($T = 1$, *clustering* = 0) is used as x-intercept; (3) from the equation of this line and the clustering coefficient of the given network, we can estimate its temperature $T$ [18]. Although this procedure is possible when few networks have to be embedded, it becomes too time consuming in wide simulations. Therefore, in all our synthetic and real mappings we used a default value $T = 0.1$. We let notice that most of this Methods section is equivalent to an analogous Methods section present in other studies of the authors [7], [13].

|           | N    | E     | m    | Cl   | γ    | C  |
|-----------|------|-------|------|------|------|----|
| karate    | 34   | 78    | 2.3  | 0.59 | 2.1  | 2  |
| opsahl 8  | 43   | 193   | 4.5  | 0.61 | 8.2  | 7  |
| opsahl 9  | 44   | 348   | 7.9  | 0.68 | 5.9  | 7  |
| opsahl 10 | 77   | 518   | 6.7  | 0.66 | 5.1  | 4  |
| opsahl 11 | 77   | 1088  | 14.1 | 0.72 | 4.9  | 4  |
| polbooks  | 105  | 441   | 4.2  | 0.49 | 2.6  | 3  |
| football  | 115  | 613   | 5.3  | 0.40 | 9.1  | 12 |
| polblogs  | 1222 | 16714 | 13.7 | 0.36 | 2.4  | 2  |

**Suppl. Table 1. Statistics of real networks.** Number of nodes $N$, number of edges $E$, half of average node degree $m$, clustering coefficient $Cl$, power-law degree distribution exponent $\gamma$, number of communities $C$.